\documentclass[%
 aip,
 amsmath,amssymb,
 preprint,%
 reprint,%
]{revtex4-1}
\DeclareUnicodeCharacter{0308}{\'{e}}
\usepackage{graphicx}
\usepackage{dcolumn}
\usepackage{bm}

\usepackage[utf8]{inputenc}
\usepackage[T1]{fontenc}
\usepackage{mathptmx}
\usepackage{etoolbox}

\usepackage{amsthm}

\usepackage[all,cmtip]{xypic}
\usepackage{placeins}
\usepackage{xcolor}

\usepackage[ruled,linesnumbered]{algorithm2e}

\makeatletter
\def\@email#1#2{%
 \endgroup
 \patchcmd{\titleblock@produce}
  {\frontmatter@RRAPformat}
  {\frontmatter@RRAPformat{\produce@RRAP{*#1\href{mailto:#2}{#2}}}\frontmatter@RRAPformat}
  {}{}
}%

\theoremstyle{plain}
\newtheorem{thm}{\protect\theoremname}
\theoremstyle{remark}
\newtheorem{rem}[thm]{\protect\remarkname}

\makeatother
\usepackage{babel}
\providecommand{\remarkname}{Remark}
\providecommand{\theoremname}{Theorem}

\begin{document}


\title[Reaction coordinate flows]{Reaction coordinate flows for model reduction of molecular kinetics}
\author{Hao Wu$^*$}
\affiliation{ 
School of Mathematical Sciences, Institute of Natural Sciences and MOE-LSC, Shanghai Jiao Tong University, Shanghai, P.~R.~China}%
\affiliation{ 
School of Mathematical Sciences, Tongji University, Shanghai, P.~R.~China}%
\author{Frank No\'e$^*$}%
\email[Authors to whom correspondence should be addressed: ]{\texttt{hwu81@sjtu.edu.cn} and \texttt{frank.noe@fu-berlin.de}}
\affiliation{ 
Department of Mathematics and Computer Science and Department of Physics, Freie Universität Berlin, Berlin, Germany}%
\affiliation{ 
Department of Chemistry, Rice University, Houston, TX, USA}%
\affiliation{ 
Microsoft Research AI4Science, Berlin, Germany}%

\begin{abstract}
In this work, we introduce a flow based machine learning approach, called reaction coordinate (RC) flow, for discovery of low-dimensional kinetic models of molecular systems.
The RC flow utilizes a normalizing flow to design the coordinate transformation and a Brownian dynamics model to approximate the kinetics of RC, where all model parameters can be estimated in a data-driven manner. In contrast to existing model reduction methods for molecular kinetics, RC flow offers a trainable and tractable model of reduced kinetics in continuous time and space due to the invertibility of the normalizing flow. Furthermore, the Brownian dynamics-based reduced kinetic model investigated in this work yields a readily discernible representation of metastable states within the phase space of the molecular system.
Numerical experiments demonstrate how effectively the proposed method discovers interpretable and accurate low-dimensional representations of given full-state kinetics from simulations.
\end{abstract}

\maketitle

\section{Introduction}
As one of the basic tasks of analysis of molecular dynamics (MD) simulations, model reduction is important for understanding the kinetics of molecular systems,
which aims to
find a small number of observables
of the molecular configuration, usually called reaction coordinate (RC),
and projects the original molecular kinetics to the space of RC.
An ideal RC is a sufficient statistics which can accurately predict the evolution of configurations in future,
so that the reduced kinetics of RC can preserve essential part of the full-state kinetics.

Because the manual selection of RCs heavily relies on prior knowledge of molecular systems and requires repeated trial-and-errors in applications, a lot of data-driven methods for model reduction have been proposed in the past decade. According to transfer operator theory, 
we can select slow variables with large time-lagged correlations as RCs
for molecular systems with fast–slow time scale separation\cite{noe2013variational,nuske2014variational,klus2016numerical,klus2018data,konovalov2021markov}. For example, the well-established time-lagged independent component analysis (TICA) method \cite{perez2013identification,schwantes2013improvements}, which is similar to the dynamic mode decomposition in the field of dynamical systems \cite{mezic2005spectral}, represents slow variables by linear transformations of the original molecular coordinates. It is shown in \cite{wu2020variational} that TICA and its nonlinear extensions \cite{boninsegna2015investigating,williams2015data,schwantes2015modeling,nuske2016variational,wu2017variational,scherer2019variational} can be considered as special cases of the variational approach for Markov processes (VAMP), and several deep learning based methods have been proposed within the VAMP framework \cite{mardt2018vampnets,sidky2019high,bonati2021deep,ghorbani2022graphvampnet}.
However, in the case where there is no significant time scale gap, it is difficult to obtain effective low-dimensional description of the molecular kinetics by collecting all slow components. In order to alleviate this problem, some model reduction methods, which directly minimize the prediction errors of future configurations (or configuration distributions) for given RCs by deep learning or kernel machines, have been proposed. Examples include the time-lagged autoencoder \cite{wehmeyer2018time}, transition manifold method \cite{bittracher2018transition,bittracher2019dimensionality}, and past-future information bottleneck method \cite{wang2019past}. Moreover, for investigation of transitions between the specific reactant and product states of a molecular system, RC can also be obtained from the committor function \cite{pozun2012optimizing,lu2014exact}.

After obtaining RC, the reduced kinetics along RC can be represented by a stochastic governing equation according to the theory of effective dynamics \cite{legoll2010effective,zhang2016effective}, but the computation and analysis of the equation are usually infeasible for high-dimensional molecular systems. Some recent modeling methods of dynamical systems can discover RC with a simple governing equation of the reduced kinetics by assuming the existence of the inverse function from RC to the system state \cite{lusch2018deep,champion2019data}. However, these methods are not applicable to MD simulations, where the assumption does not hold and the conditional distribution of the molecular configuration for given RC is generally intractable (see discussion in Section \ref{subsec:Data-driven-model-reduction}).

In this work, we introduce an innovative model reduction technique based on normalizing flow (NF), called RC flow, which enables the simultaneous discovery of nonlinear reaction coordinate transformations and the modeling of corresponding reduced kinetics.
Normalizing Flows (NFs) are a class of invertible neural network known for their capacity to perform explicit density estimation on high-dimensional data \cite{kobyzev2020normalizing}, which have also proven to be effective tools for modeling and predicting equilibrium distributions in molecular systems \cite{noe2019boltzmann,wu2020stochastic,kohler2020equivariant,kohler2021smooth,invernizzi2022skipping,dibak2022temperature,kohler2023rigid}, as well as in other many-body systems \cite{li2018neural,kanwar2020equivariant}.
Notably, in the Timewarp method recently proposed by Klein et al. \cite{klein2023timewarp}, NFs have demonstrated remarkable proficiency in constructing transferable kinetic models for molecular systems over extended timescales.
Our method utilizes an NF to decompose the full-state configuration coordinate into RC and white noise independent of kinetics, then the surrogate model of the full-state molecular kinetics can be explicitly reconstructed from the governing equation of RC due to the invertibility of NF. We model the governing equation by Brownian dynamics for interpretability and tractability of the reduced kinetics, and all parameters of the NF and Brownian dynamics can be obtained via maximum likelihood estimation from simulation data.
RC flow stands out as an ideal choice for molecular kinetics model reduction, distinguished by several unique traits:
\begin{itemize}
\item It introduces a scalable approach to model reduction that extends beyond the traditional framework of identifying slow variables. In contrast to conventional methods, which solely focus on locating slow variables, RC flow enables the collaborative description of multiple dominant components of molecular kinetics through more effective and small-sized RCs.
\item RC flow enables simultaneous training of coordinate transformations and reduced kinetics, and the model of reduced kinetics can be specified according to practical requirements.
\item The Brownian dynamics-based reduced kinetic model considered in this work can furnish a clear and low-dimensional depiction of metastable states within the phase space of the molecular system.
\end{itemize}
Several numerical tests are presented to illustrate the effectiveness of the RC flow.

\section{Background}

\subsection{Reaction coordinate\label{subsec:Data-driven-model-reduction}}
In MD simulations, the time evolution of a molecular system can be considered as a Markov process $\{\mathbf x_t\}$ in a high-dimensional configuration space contained in $\mathbb R^D$,
and the molecular kinetics can be fully described by the transition density
$p_{\tau}^{x}(\mathbf x,\mathbf y)=\mathbb{P}(\mathbf{x}_{t+\tau}=\mathbf y|\mathbf{x}_{t}=\mathbf x)$,
where $\mathbf x_t$ denotes the system configuration at time $t$.
Furthermore, we assume in this paper that the MD simulation is performed without applying any external force and the time-reversibility 
\[\mu^x(\mathbf x)p_{\tau}^{x}(\mathbf x,\mathbf y)=\mu^x(\mathbf y)p_{\tau}^{x}(\mathbf y,\mathbf x)\]
is fulfilled,
where $\mu^x(\mathbf x)$ represents the equilibrium distribution of the configuration.

If we select RC as $\mathbf{z}_{t}=\Phi(\mathbf{x}_{t})\in\mathbb{R}^{d}$
with $d\ll D$ and model the kinetics embedded in the RC space by
the transition density
$p_{\tau}^{z}(\mathbf{z}_{t},\mathbf{z}_{t+\tau})=\mathbb{P}(\mathbf{z}_{t+\tau}|\mathbf{z}_{t})$, the full-state
kinetics can be reconstructed from the reduced kinetics as
\begin{equation}
\hat{p}_{\tau}^{x}(\mathbf{x}_{t},\mathbf{x}_{t+\tau})=\int p_{\tau}^{z}(\mathbf{z}_{t},\mathbf{z})\mathbb{P}\left(\mathbf{x}_{t+\tau}|\Phi(\mathbf{x}_{t+\tau})=\mathbf{z}\right)\mathrm{d}\mathbf{z},\label{eq:reconstruction}
\end{equation}
and the relationship between $\mu^x$ and the equilibrium distribution $\mu^z$ of RC is provided by
\begin{equation}\label{eq:reconstruction-equilibrium}
\mu^{x}(\mathbf{x})=\int\mu^{z}(\mathbf{z})\mathbb{P}\left(\mathbf{x}|\Phi(\mathbf{x})=\mathbf{z}\right)\mathrm{d}\mathbf{z},
\end{equation}
where $\mathbb{P}\left(\mathbf{x}|\Phi(\mathbf{x})=\mathbf{z}\right)$
denotes the conditional distribution of the configuration $\mathbf x$ for given RC $\mathbf z$.

The relationship between the variables and transition densities is shown in the diagram below:
\begin{equation*}\label{eq:diag}
\begin{gathered}
\xymatrix@C+24pt@R+12pt{   
\mathbf{x}_t \ar[r]^{p_{\tau}^x(\mathbf x_t, \mathbf x_{t+\tau})} \ar[d]_{\Phi} & \mathbf{x}_{t+\tau} \ar[d]^{\Phi} \\   
\mathbf{z}_t \ar[r]_{p_{\tau}^z(\mathbf z_t, \mathbf z_{t+\tau})} & \mathbf{z}_{t+\tau}
}
\end{gathered}
\end{equation*}
It can be observed from the diagram that
model parameters of the coordinate transformation $\Phi$ and the reduced transition density
$p_{\tau}^{z}$ can be fitted by maximizing the likelihood of $\hat{p}_{\tau}^{x}$ for given simulation trajectories.
A major difficulty of such a model reduction procedure arises from the density 
$\mathbb{P}\left(\mathbf{x}|\Phi(\mathbf{x})=\mathbf{z}\right)$ in (\ref{eq:reconstruction},\ref{eq:reconstruction-equilibrium}),
which is a probability distribution over the level set $\left\{\mathbf x\vert\Phi(\mathbf{x})=\mathbf{z}\right\}$ of $\Phi$
and is generally intractable for conventional probabilistic models
especially for nonlinear $\Phi$.
In what follows, we will address this difficulty by using normalizing flows.

\subsection{Normalizing flow}

Normalizing flows (NFs) are a specific class of neural networks for
modeling distributions of high-dimensional data \cite{kobyzev2020normalizing,papamakarios2021normalizing}. For a random variable $\mathbf{x}$,
we can train an NF $F$ so that $\bar{\mathbf z} = F(\mathbf{x})$ is distributed according
to a tractable distribution, e.g., standard Gaussian distribution $\mathcal N(\mathbf 0, \mathbf I)$,
where $F$ is a bijective function consisting of a sequence of invertible transformations.
After training, we can approximate the density of $\mathbf{x}$ as
\begin{equation}\label{eq:nf-transformation}
\mathbb{P}(\mathbf{x})=\mathcal{N}\left(\bar{\mathbf z}|\mathbf{0},\mathbf{I}\right)\left|\mathrm{det}\left(\frac{\partial F(\mathbf{x})}{\partial\mathbf{x}}\right)\right|
\end{equation}
according to the change of variables formula and draw samples from the density by
\begin{eqnarray}
\bar{\mathbf{z}} & \sim & \mathcal{N}(\mathbf{0},\mathbf{I}),\nonumber \\
\mathbf{x} & = & F^{-1}(\bar{\mathbf{z}}),\label{eq:nf-sample}
\end{eqnarray}
where $\mathcal{N}\left(\cdot|\mathbf{a},\boldsymbol{\Sigma}\right)$ denotes the Gaussian density function
with mean $\mathbf a$ and covariance matrix $\boldsymbol{\Sigma}$.
There have been many types of NF models available in literature \cite{dinh2014nice,rezende2015variational,dinh2016density,grathwohl2018ffjord,tang2020deep}. For most models, the Jacobian determinant of
$F$ and the inverse $F^{-1}$
in Eqs.~(\ref{eq:nf-transformation}, \ref{eq:nf-sample}) can be efficiently calculated due to the specifically designed network structures.

\section{Reaction coordinate flow}
\subsection{Model architecture}
\begin{figure}
\begin{centering}
\includegraphics[width=1\columnwidth]{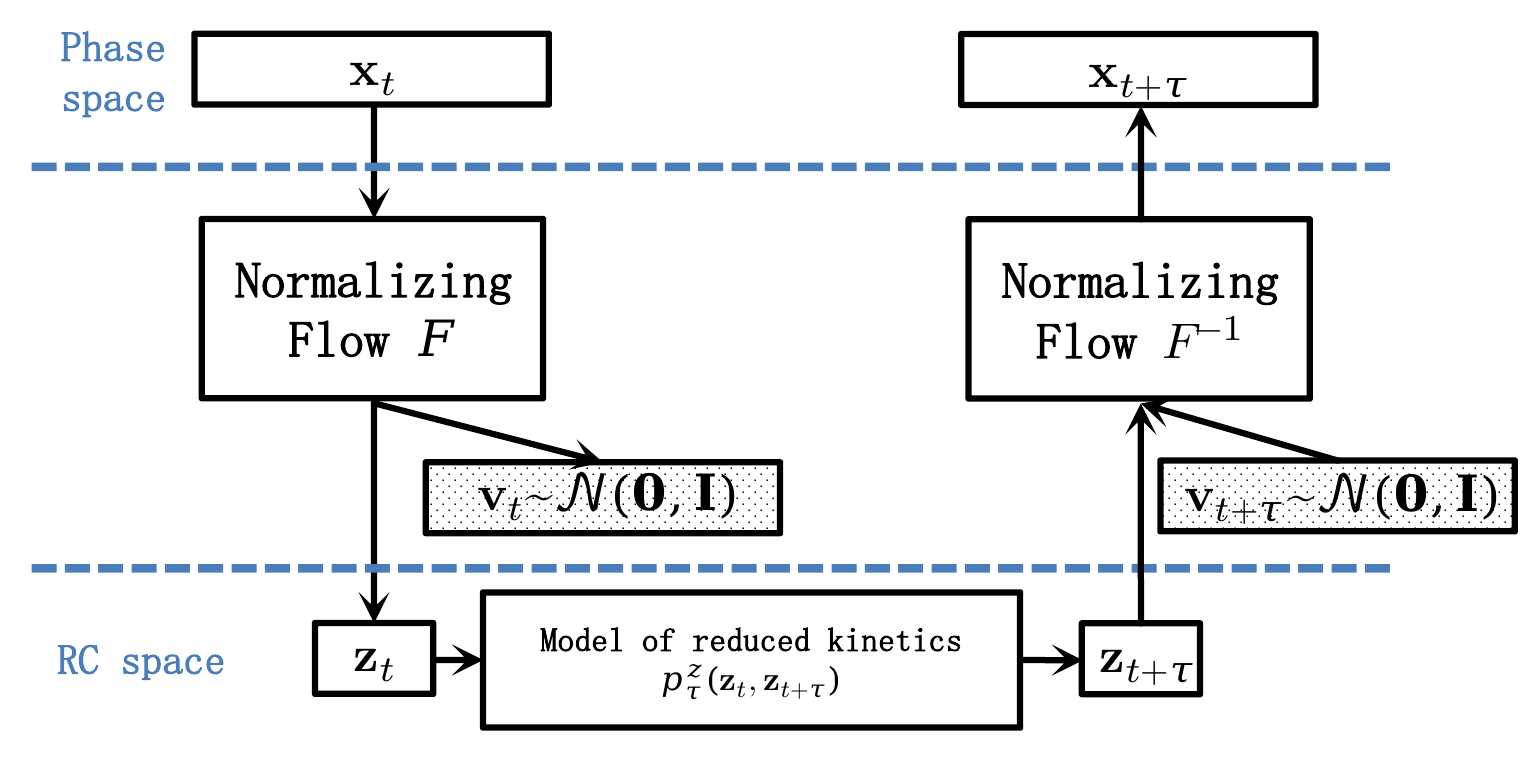}
\par\end{centering}
\caption{Diagram of RC flow for model reduction of molecular kinetics. The NF $F$ transforms state $\mathbf{x}_{t}$ into a vector consisting
of the RC $\mathbf{z}_{t}$ and the noise $\mathbf{v}_{t}\sim\mathcal{N}(\mathbf{0},\mathbf{I})$.
Correspondingly, $\mathbf{x}_{t+\tau}$ can be reconstructed from
$\mathbf{z}_{t+\tau}$ as $\mathbf{x}_{t+\tau}=F^{-1}(\mathbf{z}_{t+\tau},\mathbf{v}_{t+\tau})$
by sampling $\mathbf{v}_{t+\tau}\sim\mathcal{N}(\mathbf{0},\mathbf{I})$.
The time evolution of $\mathbf{z}_{t}$ is characterized by Brownian dynamics \eqref{eq:continuous-brownian} in this paper.\label{fig:Diagram-of-kinetic-flow}}
\end{figure}

In this section, we develop a RC flow for model reduction of MD simulation data, which can identify low-dimensional RC and model reduced kinetics simultaneously. The key idea is to utilize an NF $F$ to decompose the configuration $\mathbf{x}_{t}$ into RC and noise parts as the following:
\begin{equation}\label{eq:rcf-decomposition}
F(\mathbf{x}_{t}) = (\mathbf{z}_{t},\mathbf{v}_{t}),
\end{equation}
where the RC $\mathbf{z}_{t}$ is governed by a low-dimensional kinetic model and
the noise
$\mathbf{v}_{t}\stackrel{\mathrm{iid}}{\sim}\mathcal{N}(\mathbf{0},\mathbf{I})$ is uninformative in predicting future evolution of configurations. A diagram illustrating the model architecture is shown in Fig.~\ref{fig:Diagram-of-kinetic-flow}.

From the decomposition \eqref{eq:rcf-decomposition} and the invertibility of $F$, we can obtain the coordinate transformation
\[
\Phi(\mathbf{x}) = \left[F(\mathbf{x})\right]_{1,\ldots,d}
\]
and an analytical expression of the conditional distribution of $\mathbf x$ over the $\mathbf z$-level set
\begin{eqnarray}
\mathbb{P}\left(\mathbf{x}|\Phi(\mathbf{x})=\mathbf{z}\right) & = & \mathbb{P}\left(\mathbf{z},\mathbf{v}|\Phi(\mathbf{x})=\mathbf{z}\right)\left|\mathrm{det}\left(\frac{\partial F(\mathbf{x})}{\partial\mathbf{x}}\right)\right|\nonumber\\
 & = & \delta_{\mathbf z}\left(\Phi(\mathbf{x})\right)S(\mathbf{x}),\label{eq:decoder}
\end{eqnarray}
where
\[
S(\mathbf{x})=\mathcal{N}(\mathbf{v}|\mathbf{0},\mathbf{I})\left|\mathrm{det}\left(\frac{\partial F(\mathbf{x})}{\partial\mathbf{x}}\right)\right|
\]
and $\delta_{\mathbf z}$ denotes the Dirac delta distribution centered at $\mathbf z$.
By substituting \eqref{eq:decoder} into (\ref{eq:reconstruction},\ref{eq:reconstruction-equilibrium}), we can recover the full-state thermodynamics and kinetics from the reduced ones as
\begin{eqnarray}
\hat{p}_{\tau}^{x}(\mathbf{x}_{t},\mathbf{x}_{t+\tau}) & = & p_{\tau}^{z}(\Phi(\mathbf{x}_{t}),\Phi(\mathbf{x}_{t+\tau}))S(\mathbf{x}_{t+\tau})\label{eq:reconstruction-rcf}\\
\mu^{x}(\mathbf{x}) & = & \mu^{z}(\Phi(\mathbf{x}))S(\mathbf{x})\label{eq:equilibrium-rcf}
\end{eqnarray}

In RC flow, the reduced kinetics $p_{\tau}^{z}$ can be modeled by any
kind of kinetic model according to the practical requirements.
In this paper, we focus on the Brownian dynamics
\begin{equation}\label{eq:continuous-brownian}
\mathrm{d}\mathbf{z}_{t}=-\nabla V(\mathbf{z}_{t})\mathrm{d}t+\sqrt{2\beta^{-1}}\mathrm{d}W_{t},
\end{equation}
which is widely used in computational chemistry and physics for characterizing diffusion processes in complicated energy landscapes \cite{ren2002probing,singhal2004using,anderson2013compartmental}.
Here $W_t$ denotes standard Brownian motion, $V$ is potential function,
$\beta$ is inverse temperature, which we take to be $1$ without loss of generality (see Appendix \ref{app:beta}),
and the asymptotic steady-state probability distribution of $\mathbf z_t$ is
\cite{risken1996fokker}
\begin{equation}\label{eq:equilibrium-z}
\mu^z(\mathbf z)\propto \exp\left({-V(\mathbf z)}\right).
\end{equation}
The closed-form of the transition density $p_{\tau}^z$ of the Brownian dynamics \eqref{eq:continuous-brownian} is generally unavailable,
but there are many numerical methods that can produce numerical approximations effectively (see survey in \cite{jensen2002transition}).
In our experiments, we select the importance sampling method proposed in \cite{durham2002numerical}, which is outlined in Appendix \ref{app:pz}.
\begin{rem}
Conditional NF can also be used to model $p_\tau^z$, as demonstrated in \cite{klein2023timewarp}, avoiding the complex numerical computations associated with SDEs. Nevertheless, this approach sacrifices interpretability in reduced kinetics and presents challenges in obtaining the reduced potential of $\mathbf z_t$.
\end{rem}

Here, we model $F$ by RealNVP \cite{dinh2016density}. For $\mu^z$, which is a low-dimensional density and usually multimodal, we select Gaussian mixture model (GMM) in the form of
\begin{equation}\label{eq:gmm}
\hat{\mu}^{z}(\mathbf{z})=\sum_{i=1}^{K^d}\frac{w(\mathbf{c}_{i})}{\sum_{j=1}^{K^d}w(\mathbf{c}_{j})}\mathcal{N}\left(\mathbf{z}|\mathbf{c}_{i},\mathrm{diag}\left(\boldsymbol{\sigma}(\mathbf{c}_{i})\right)^2\right)
\end{equation}
according to our numerical experience. In \eqref{eq:gmm}, centers $\mathbf c_1,\ldots,\mathbf c_{K^d}$ are kept as grid points of a regular grid in the $d$-dimensional RC space (see Line \ref{alg-line:c} in Algorithm \ref{alg:rc}), and $w(\mathbf c)\in \mathbb R, \boldsymbol{\sigma}(\mathbf c)\in \mathbb R^d$ are considered as functions of centers, which are both represented by multilayer perceptrons with exponential output activation functions. The drift term in the reduced kinetics \eqref{eq:continuous-brownian} can then be calculated as
\[
-\nabla V=\nabla \log \mu^z.
\]

\begin{rem}
RC flow can also be characterized by parametric models of $F$ and $V$. But in this case, the calculation of
\[
\hat{\mu}^{x}(\mathbf{x})=\mu^{z}(\Phi(x))S(\mathbf{x})=\frac{\exp\left(-V(\Phi(\mathbf{x}))\right)S(\mathbf{x})}{\int\exp\left(-V(\mathbf{z})\right)\mathrm{d}\mathbf{z}}
\]
involves an intractable integral over the RC space.
\end{rem}

\subsection{Loss functions}
In order to identify RC that can characterize thermodynamics and kinetics of the molecular system,
we consider the following two loss functions for training of RC flow:
\begin{description}
\item [{Kinetic loss}] To recover full-state kinetics from the RC flow, we use the negative log-likelihood (NLL) loss of the estimated transition density
\begin{equation}\label{eq:loss-kin}
\mathcal L_{\mathrm{kin}}=-\frac{1}{T-\tau}\sum_{t=1}^{T-\tau}\log \hat{p}_{\tau}^{x}(\mathbf{x}_{t},\mathbf{x}_{t+\tau})
\end{equation}
for a given simulation trajectory $\{\mathbf x_1,\ldots,\mathbf x_T\}$, where $\hat{p}_{\tau}^{x}$ is defined by \eqref{eq:reconstruction-rcf}.
In the case of multiple trajectories, $\mathcal L_{\mathrm{kin}}$ can be defined as the average NLL over all transition pairs with lag time $\tau$ within all trajectories.
\item [{Equilibrium loss}] An ideal RC flow also allows an accurate estimation of the steady-state
probability distribution, and the estimation error can be achieved using the loss
\begin{equation}\label{eq:loss-eq}
\mathcal{L}_{\mathrm{eq}}=-\frac{1}{T^{s}}\sum_{t=1}^{T^{s}}\log\hat{\mu}^{x}(\mathbf{x}_{t}^{s}),
\end{equation}
where $\hat{\mu}^{x}$ is given by \eqref{eq:equilibrium-rcf} and $\{\mathbf{x}_{1}^{s},\ldots,\mathbf{x}_{T^{s}}^{s}\}$ are sampled from the global equilibrium (e.g., from enhanced sampling simulations). In numerical experiments of this paper, MD simulation trajectories are long enough and achieve equilibrium, so we simply set $\{\mathbf{x}_{1}^{s},\ldots,\mathbf{x}_{T^{s}}^{s}\}=\{\mathbf x_1,\ldots,\mathbf x_T\}$.
\end{description}

Therefore, leaving aside numerical details, we can solve the following problem with weight $\alpha>0$ to find optimal parameters of the RC flow:
\begin{equation}\label{eq:loss}
\min_{F, \mu^z} \mathcal L = \mathcal L_{\mathrm{kin}} + \alpha \mathcal{L}_{\mathrm{eq}},
\end{equation}

\subsection{Pre-processing and training}
It is well-known that the performance of training neural networks can be significantly improved by normalizing the data scales and decoupling the correlation between features of the data. In a previous study \cite{noe2019boltzmann}, it was also reported that whitening transformation can enhance the performance of learning Boltzmann distributions by normalizing flows (NFs). In this paper, we perform data pre-processing by TICA\cite{perez2013identification,schwantes2013improvements} for large-sized systems and let
$\mathbf x_t^\mathrm{TICA} = \mathbf W(\mathbf x_t-\bar{\mathbf x})$,
where $\bar {\mathbf x}$ denotes the mean value of the original configuration and $W$ denotes the transformation matrix given by TICA.
If the features of the original configuration $x_t$ are linearly independent, then $W$ is invertible, and $F(\mathbf W(\mathbf x-\bar{\mathbf x}))$ is still an invertible function with respect to $\mathbf x$. This data pre-processing approach offers several advantages. First, the features of $\mathbf x_t^\mathrm{TICA}$ are orthonormal with an identity covariance matrix, making them nearly standard Gaussian distributed. Second, the TICA components are sorted according to their relaxation time scales, with the slowest and most important components arranged in the first dimensions. Therefore, $\mathbf x_t^\mathrm{TICA}$ is already close to an ideal combination of reaction coordinate and noise after the TICA transformation, which simplifies the learning problem. Further details on the TICA can be found in Appendix \ref{app:tica}.

Based on the above analysis, Algorithm \ref{alg:rc} provides a summary of the training process of RC flow. In the algorithm, we first simplify the reduced kinetics as a Brownian motion with $\nabla V(\mathbf z)\equiv 0$ and initialize $F$ by minimizing $\mathcal L_{\mathrm{kin}}$. Next, we obtain a rectangle area $\prod_{i=1}^{d}[\mathrm{LB}_{i},\mathrm{UB}_{i}]$ that covers all RCs of training data, locate the $K^d$ centers of the GMM of $\mu^z$ uniformly in the area, and initialize the weights and variance matrices by minimizing $\mathcal L$. Last, all parameters of $F$ and $\mu^z$ are simultaneously optimized with objective function $\mathcal L$. In our experiments, all the optimization problems involved in the algorithm are solved by the mini-batch Adam algorithm \cite{kingma2014adam}.

Notice the pre-processing and pre-training steps are included in the algorithm to stabilize the training, and can be replaced with other heuristic strategies according to practical requirements.

\begin{rem}
In this paper, the transformation coefficients $\mathbf W, \bar{\mathbf x}$ obtained by TICA remain unchanged during the training procedure.
It is important to note that some recently proposed normalizing flow models (e.g., KRnet \cite{tang2020deep}) incorporate trainable linear layers with invertible transformation matrices. The application of such NFs to the RC flow framework requires further investigation and evaluation.
\end{rem}

\begin{algorithm}
  \caption{Training algorithm of RC flow}
  \label{alg:rc}
  \SetAlgoNoLine
   \If{\rm{data pre-processing is required}}{
   Perform TICA transformation
   \begin{equation*}
   \mathbf x_t := \mathbf W(\mathbf x_t-\bar{\mathbf x}) \text{ for all $t$.}
   \end{equation*}}
   \If{\rm{pre-training is required}}{
   Perform pre-training (optional): \\
   Let $\mu^z(\mathbf z)\propto 1$, i.e., $\nabla V(\mathbf z)\equiv 0$, and train $F$ with $\mathcal L_{\mathrm{kin}}$ defined in \eqref{eq:loss-kin}. \\
   Calculate $(\mathbf z_t,\mathbf v_t)=F(\mathbf x_t)$ for $t=1,\ldots,T$. \\
   Let
\begin{eqnarray*}
\mathrm{LB}_{i} & = & \min\{i\text{th element of }\mathbf{z}_{t}|t=1,\ldots,T\},\\
\mathrm{UB}_{i} & = & \max\{i\text{th element of }\mathbf{z}_{t}|t=1,\ldots,T\}.
\end{eqnarray*}
for $i=1,\ldots,d$. \\
   Let
\[
\{\mathbf{c}_{1},\ldots,\mathbf{c}_{K^{d}}\}=\prod_{i=1}^{d}\left\{ \mathrm{LB}_{i}+\frac{k\left(\mathrm{UB}_{i}-\mathrm{LB}_{i}\right)}{K-1}|k=0,\ldots,K-1\right\} 
\]\label{alg-line:c} \\
   Train $w(\mathbf c), \boldsymbol{\sigma}(\mathbf c)$ with $\mathcal L$ defined in \eqref{eq:loss} while keeping $F$ fixed.\label{alg-line:wsigma}}
   Train $F$ and $w(\mathbf c), \boldsymbol{\sigma}(\mathbf c)$ simultaneously with $\mathcal L$. \\
   \textbf{Return} $F(\cdot)$ (or $F(\mathbf W(\cdot-\bar{\mathbf x}))$ if Line 2 is implemented).
\end{algorithm}
\section{Analysis}
\subsection{Spectral analysis}
It is well known that the Brownian dynamics \eqref{eq:continuous-brownian} is a time-reversible Markov process and its transition density can be decomposed into a set of relaxation processes as \cite{noe2013variational,nuske2014variational}
\begin{equation}\label{eq:eigendecomposition-z}
p_\tau^z(\mathbf{z}_t,\mathbf{z}_{t+\tau})=\sum_{i=0}^{\infty}r_i^z(\mathbf z_t)\cdot e^{-\frac{\tau}{t_i}}r_i^z(\mathbf z_{t+\tau})\mu^z(\mathbf z_{t+\tau}),
\end{equation}
for all $\tau>0$, where $t_0=\infty>t_1\ge t_2\ge\ldots$ are implied time scales, $r_0^z,r_1^z,,r_2^z\ldots$ are eigenfunctions of the transfer operator of $\{\mathbf z_t\}$, and $r_0^z(\mathbf z)\equiv 1$. If there is a spectral gap and $t_1\ge \ldots \ge t_{s-1}\gg t_s$, we can conclude that the kinetics of
$\{\mathbf z_t\}$ at large time scales is dominated by the first $s$ relaxation processes.

In the case where the RC flow is well trained
and provides an accurate approximation of $p_\tau^x$,
it can be proved that $t_0,t_1,t_2\ldots$ are also implied time scales of $\{\mathbf{x}_t\}$, and we can obtain the dominant eigenfunctions and relaxation processes of $\{\mathbf x_t\}$ from those of RC with
\begin{equation}
r_i^x(\mathbf x)=r_i^z(\Phi(\mathbf x))
\end{equation}
and
\begin{equation}\label{eq:eigendecomposition-x}
p_\tau^x(\mathbf{x}_{t},\mathbf{x}_{t+\tau})=\sum_{i=0}^{\infty}r_{i}^{x}(\mathbf{x}_t)\cdot e^{-\frac{\tau}{t_{i}}}r_{i}^{x}(\mathbf{x_{t+\tau}})\mu^{x}(\mathbf{x}_{t+\tau}).
\end{equation}
The detailed derivations of the above conclusions are given in Appendix \ref{app:transfer-operator}.

\subsection{RC flow and information bottleneck}
One important metric to evaluate the quality of RC is the mutual information $\mathrm{MI}(\mathbf z_t,\mathbf x_{t+\tau})$, which quantifies the statistical dependence between $\mathbf z_t$ and $\mathbf x_{t+\tau}$. According to the principle of past–future information bottleneck \cite{still2014information,wang2019past}, if $\mathbf z_t$ is an ideal bottleneck variable with large mutual information, it can accurately predict the future evolution of the configuration.

For RC flow,  it can be established that, as the size of simulation data tends towards infinity, the following inequality holds:
\begin{equation}\label{eq:mi-bound}
\mathrm{MI}(\mathbf z_t,\mathbf x_{t+\tau}) \ge -\mathcal L_{\mathrm{kin}} + \text{const.},
\end{equation}
where the constant is independent of our model parameters.
Detailed proof of this result is provided in Appendix \ref{sec:mi-bound-proof}.
Thus, RC flow can also be interpreted as an information bottleneck method, which maximizes a lower bound of the mutual information $\mathrm{MI}(\mathbf z_t,\mathbf x_{t+\tau})$ with a specific kinetic model of $\mathbf z_t$.

\section{Numerical examples}
In this section, we apply RC flow to model reduction of some diffusion processes with multiple metastable states and the alanine dipeptide, where $F$ is modeled by a realNVP. All the model and implementation details are presented in Appendix \ref{app:details}.

\begin{figure}
\begin{centering}
\includegraphics[width=1\linewidth]{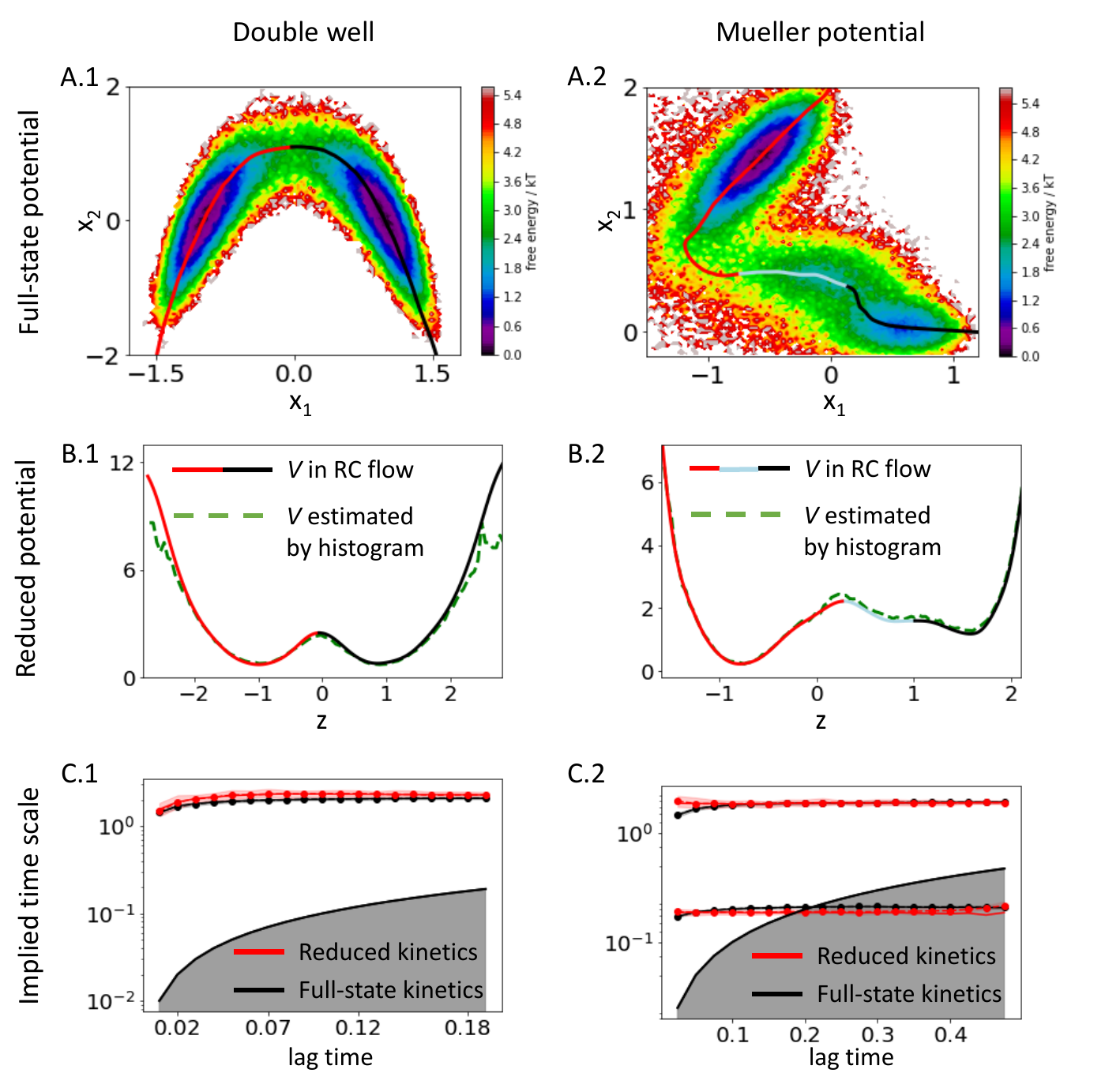}
\par\end{centering}
\caption{Model reduction of two-dimensional diffusion processes.
\textbf{(A)} Potential energy functions in $\mathbb R^2$. The solid lines consist of points $\{F^{-1}(z,0)\vert z\in\mathbb R\}$, and points with the same colors indicate they belong to the same potential well of the reduced potential $V$.
\textbf{(B)} Reduced potentials of RCs. The solid lines represent $V$ estimated by the GMM in RC flow, 
where the potential wells are created by local maxima of $V$, and the dash lines represent potentials estimated by the histogram of $\{z_t\}$.
\textbf{(C)} Dominant implied time scales of the reduced kinetics \eqref{eq:continuous-brownian} of $z_t$ (red) and the full-state Brownian dynamics of $\mathbf x_t$ (black), which are both calculated from simulation trajectories by the MSM approach at different lag times (see Appendix \ref{app:its} for details of calculation).
\label{fig:toy-example-part1}}
\end{figure}

First, we consider Brownian dynamics driven by the double well potential and the Mueller potential as shown in Fig.~\ref{fig:toy-example-part1}A, and use RC flow to identify the one-dimensional RC $z$ based on simulation trajectories in the configuration space of $\mathbf x = (x_1,x_2)$. Fig.~\ref{fig:toy-example-part1}B plots the reduced potential $V$ of $z$, which can be divided into several potential wells by barriers. We also present in Fig.~\ref{fig:toy-example-part1}A full-state configurations belonging to different potential wells through the inverse mapping $\mathbf x=F^{-1}(z,0)$ for $z\in\mathbb R$. It can be seen that RCs identified by the RC flow can preserve the structure of metastable states, and the stationary distributions of RCs are accurately estimated. In addition, we utilize Markov state models (MSMs) \cite{prinz2011markov,husic2018markov} to estimate the dominant implied time scales of the processes from the original simulation trajectories of $\mathbf x$ and trajectories of RC generated by \eqref{eq:continuous-brownian}. The results are given in Fig.~\ref{fig:toy-example-part1}C, which demonstrate the consistency between the full-state and reduced kinetics.

\begin{figure}
\begin{centering}
\includegraphics[width=1\linewidth]{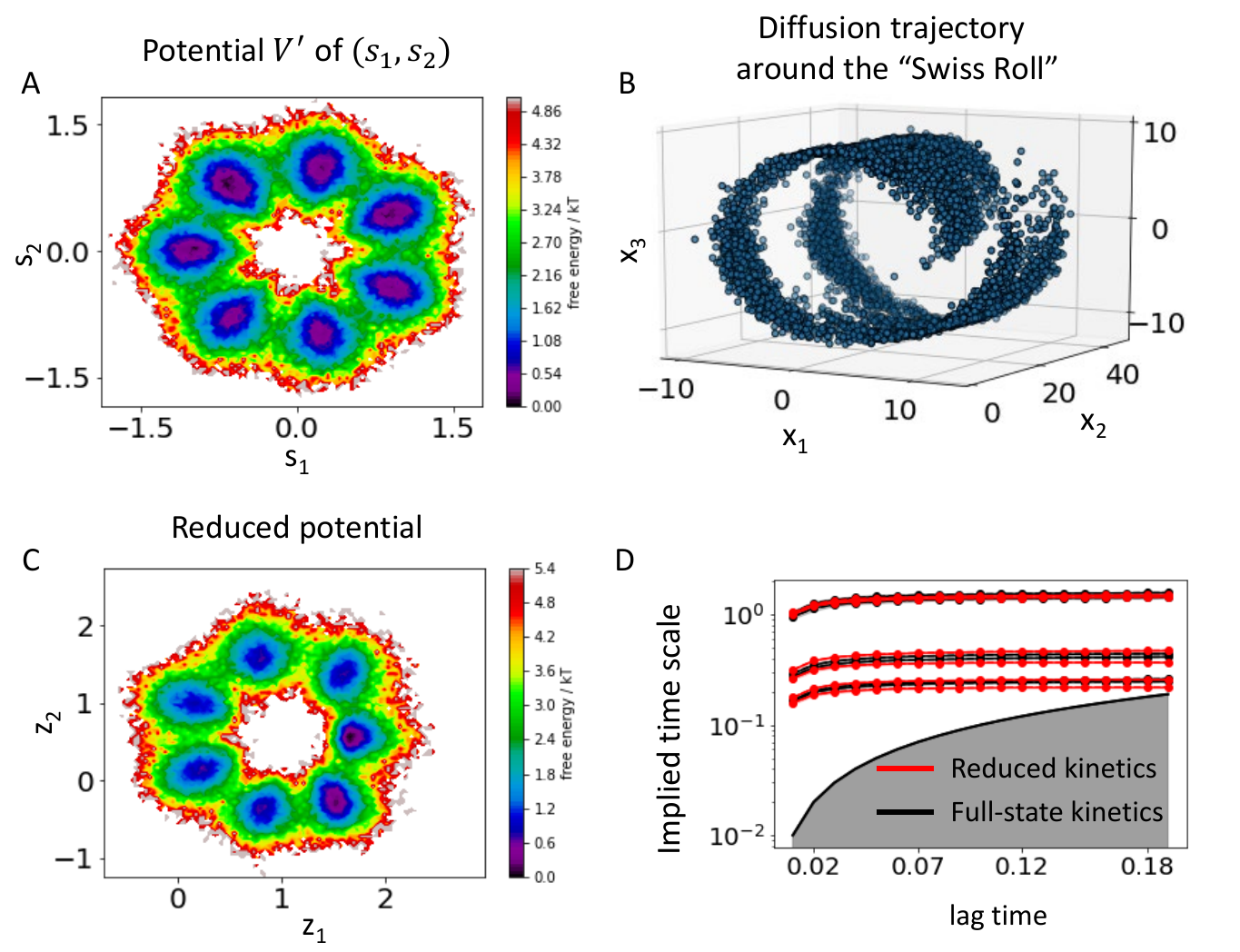}
\par\end{centering}
\caption{Model reduction of the diffusion process around a ``Swiss roll''. 
\textbf{(A)} Circular potential $V'$ of $(s_1,s_2)$ (see Appendix \ref{app:simulations}).
\textbf{(B)} Diffusion trajectory of $\mathbf x=(x_1,x_2,x_3)$ obtained by the transformation \eqref{eq:swiss-transform}.
\textbf{(C)} Reduced potential of $\mathbf z=(z_1,z_2)$ obtained by RC flow.
\textbf{(D)} Dominant implied time scales $t_1,\ldots,t_6$ of the reduced kinetics \eqref{eq:continuous-brownian} of $\mathbf z_t$ (red) and the full-state Brownian dynamics of $\mathbf x_t$ (black), which are calculated by the MSM approach at different lag times.
\label{fig:toy-example-part2}}
\end{figure}

As a second example, the data are generated by a diffusion process with potential
\[
V^s(\mathbf s)=V'(s_1,s_2)+10s_3^2,
\]
where $\mathbf s=(s_1,s_2,s_3)$ is the state, the potential $V'$ of $(s_1,s_2)$ plotted in Fig.~\ref{fig:toy-example-part2}A has seven wells, and $s_3$ evolves as an independent Ornstein-Uhlenbeck process with equilibrium distribution $\mathcal N(s_3|0,0.05)$ and the mixing time much smaller than that of $(s_1,s_2)$. In this example, the true simulation data of $\mathbf s$ are mapped to $\mathbf x=(x_1,x_2,x_3)$ lying around a ``Swiss roll'' manifold by a nonlinear transformation as shown in Fig.~\ref{fig:toy-example-part2}B, and the RC flow is implemented to find a two-dimensional RC. Figs.~\ref{fig:toy-example-part2}C and \ref{fig:toy-example-part2}D reveal that the proposed method successfully ``unfolds'' the manifold of simulation data and the kinetic properties of the process are preserved under the model reduction.

\begin{figure}
\begin{centering}
\includegraphics[width=0.5\textwidth]{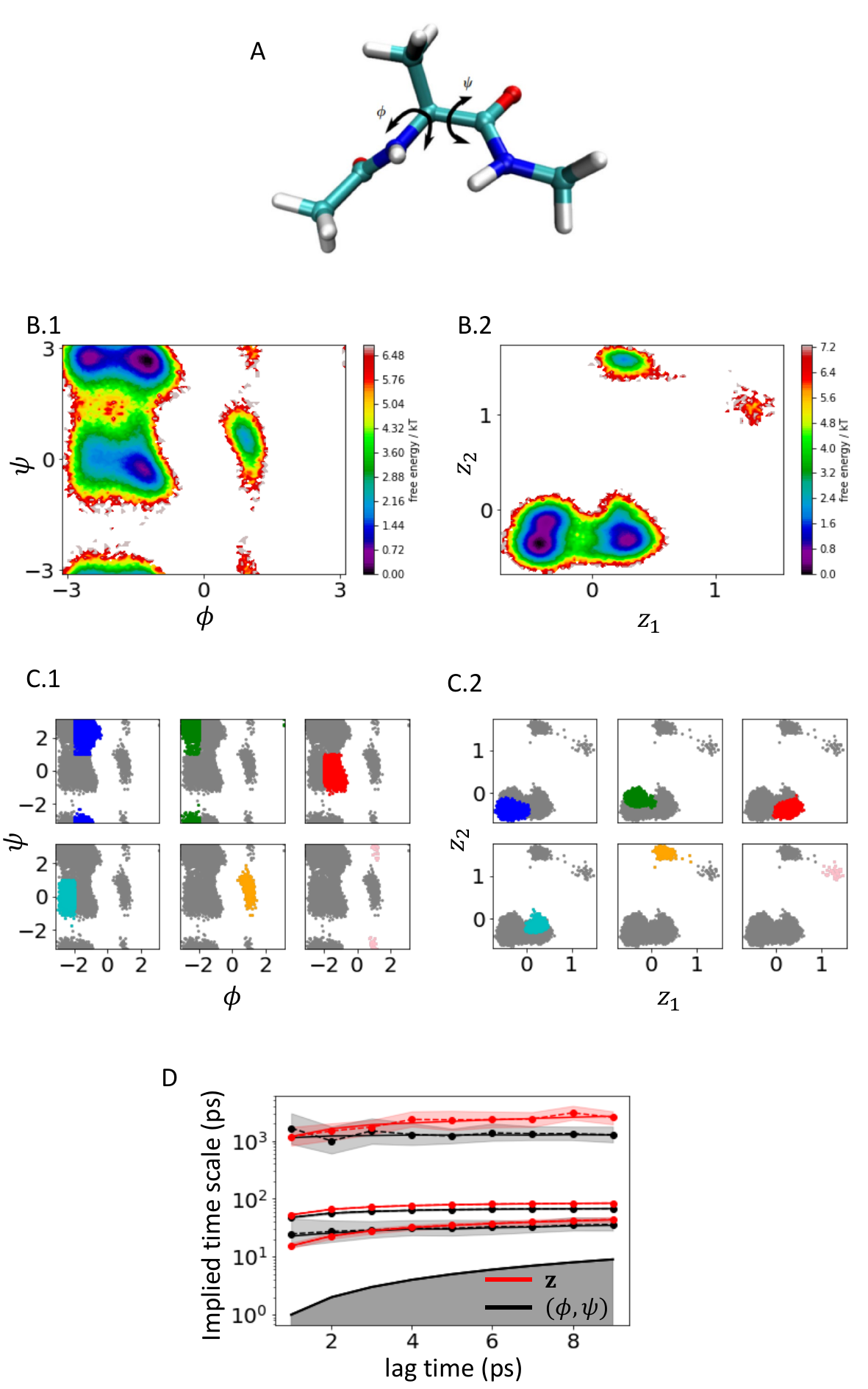}
\par\end{centering}
\caption{Model reduction of alanine dipeptide. \textbf{(A)} Structure
of alanine dipeptide. The main coordinates are the backbone torsion
angles $\phi$ and $\psi$. 
\textbf{(B)} Reduced potential of $(\phi,\psi)$ and $V(\mathbf z)$ given by the RC flow.
\textbf{(C)} Macro-states defined in the space of $(\phi,\psi)$ and their projections in the space of
$\mathbf z$. \textbf{(D)} Implied timescales $t_1,t_2,t_3$ of the reduced kinetics provided by the RC flow (red), and those estimated from
MD simulation trajectories in the space of $(\phi,\psi)$ (black).\label{fig:alanine}}
\end{figure}

Last, we use RC flow to analyze simulation data of alanine dipeptide. This molecular system has been extensively studied in the existing literature, and its kinetics on long timescales can be characterized by two dihedral angles $\phi,\psi$ of the backbone (Fig.~\ref{fig:alanine}A). Here we perform model reduction to find a two-dimensional RC with the configuration $\mathbf x\in \mathbb R^{30}$ being Cartesian coordinates of heavy atoms. From Fig.~\ref{fig:alanine}B, we can see that the free energy landscape of dihedral angles $(\phi,\psi)$ is close to that of $\mathbf z$ identified by RC flow.
We further grouped molecular configurations into six macro-states based on potential wells of $(\phi,\psi)$, and the macro-states can also be separated in the space of $\mathbf z$ (Fig.~\ref{fig:alanine}C). In addition, Fig.~\ref{fig:alanine}D shows that the largest three implied time scales of the reduced kinetics provided by RC flow are close to those calculated from MSMs in the space of $(\phi,\psi)$.

\section{Conclusions}
Model reduction of molecular kinetics, including discovery of RC and projection of kinetics into RC space, is an important task in analyzing and simulating molecular systems, and RC flow proposed in this paper provides a novel data-driven approach to the task. By harnessing the invertibility of NF, RC flow makes it tractable to compute conditional distributions of configurations based on specified RCs. Consequently, the reconstruction of full-state thermodynamics and kinetics from the reduced model becomes a straightforward process. Remarkably, within the RC flow framework, the optimization of coordinate transformations and the modeling of reduced kinetics can be simultaneously executed. Furthermore, it offers the flexibility to select from various governing equations for RCs in alignment with practical requirements.

In future, we will focus on the applications of RC flows to adaptive sampling and transition path finding methods, and the following open problems require further investigations:
\begin{itemize}
\item A rigorous mathematical analysis of modeling errors of RC flows under some proper assumptions (e.g., existence of low-dimensional transition manifold \cite{bittracher2018transition}) would be desirable.
\item In common model reduction methods, large lag times $\tau$ are usually selected to achieve effective low-dimensional description of kinetics especially for long time scales. But for RC flows, the accurate and efficient calculation of the transition density $p_\tau^z(\mathbf z_t,\mathbf z_{t+\tau})$ of the reduced kinetics with a large $\tau$ is still challenging.
\end{itemize}

\FloatBarrier

\appendix

\section{Analysis of the inverse temperature}\label{app:beta}
For a RC flow given by \eqref{eq:rcf-decomposition} and \eqref{eq:continuous-brownian} with $\beta\neq 1$,
we can define a new RC $\mathbf z^{\prime}_t$ as
\begin{equation}\label{eq:z-new}
F^{\prime}(\mathbf{x}_{t})=(\mathbf{z}_{t}^{\prime},\mathbf{v}_{t})=(\sqrt{\beta}\mathbf{z}_{t},\mathbf{v}_{t}),
\end{equation}
where $F'$ is also an invertible function.
Substituting \eqref{eq:z-new} into \eqref{eq:continuous-brownian} yields
\begin{equation}\label{eq:continuous-brownian-new}
\mathrm{d}\mathbf{z}_{t}^{\prime}=-\nabla V'(\mathbf{z}'_{t})\mathrm{d}t+\sqrt{2}\mathrm{d}W_{t}
\end{equation}
with
\[
V'(\mathbf{z}')=\beta V(\sqrt{\beta^{-1}}\mathbf{z}^{\prime}).
\]
It can be seen that (\ref{eq:z-new}, \ref{eq:continuous-brownian-new}) provide an equivalent RC flow with the inverse temperature of the reduced kinetics being $1$.

\section{Data pre-processing}\label{app:tica}
For completeness, we introduce the TICA-based data pre-processing approach. The interested readers can refer to \cite{perez2013identification, schwantes2013improvements} for further discussion and implementation guidance.

Suppose we are given a trajectory $\mathbf x_1,\ldots,\mathbf x_T$ of length $T$ and lag time $\Delta t$. For notational simplicity, we denote the mean value of $\mathbf x_t$ by $\bar{\mathbf x}$, the difference between $\mathbf x_t$ and $\bar{\mathbf x}$ by $\delta \mathbf x_t=\mathbf x_t-\bar{\mathbf x}$, and the projection of $\mathbf x_t$ onto the TICA space by $\mathbf x_t^{\mathrm{TICA}}$. TICA solves the following optimization problems iteratively for $i=1,2,\ldots$,
\[
\max_{\mathbf w_i}{\mathrm{auto}({\mathbf w_i,\Delta t})}=\mathbf w_i^\top \mathbf C(\Delta t)\mathbf w_i
\]
subject to constraints $\mathbf w_i^\top \mathbf C(0)\mathbf w_i=1$ and $\mathbf w_i^\top \mathbf C(0)\mathbf w_j=0$ for $j=1,\ldots,i-1$. Here, $\mathbf{C}(0)$ and $\mathbf{C}(\Delta t)$ are the estimated covariance and time-lagged covariance matrices of $\mathbf x_t$ given by
\begin{eqnarray*}
\mathbf{C}(0) & = & \frac{1}{T}\sum_{t}\delta\mathbf{x}_{t}\delta\mathbf{x}_{t}^{\top},\\
\mathbf{C}(\Delta t) & = & \frac{1}{2\left(T-\Delta t\right)}\sum_{t}\left(\delta\mathbf{x}_{t}\delta\mathbf{x}_{t+\Delta t}^{\top}+\delta\mathbf{x}_{t+\Delta t}\delta\mathbf{x}_{t}^{\top}\right).
\end{eqnarray*}
$\mathrm{auto}(\mathbf w_i,\Delta t)$ denotes the autocorrelation of the component $\mathbf w_i^\top \mathbf x_t$ with lag time $\Delta t$, and the largest autocorrelations yield the slowest components. By combining all optimal $\mathbf w_i$, we can obtain the TICA transformation
\[
\mathbf x_t^{\mathrm{TICA}} = \mathbf W(\mathbf x_t-\bar{\mathbf x})
\]
with $\mathbf W=(\mathbf w_1,\mathbf w_2,\ldots)^\top$, which ensures that $\mathbf x_t^{\mathrm{TICA}}$ has zero mean and identity covariance. In this paper, we use the function \texttt{pyemma.coordinates.tica} in the Python package \texttt{PyEMMA}\cite{scherer_pyemma_2015} to implement the TICA transformation.

In the case where $\mathbf C(0)$ is full-rank, $\mathbf W\in \mathbb R^{D\times D}$ is an invertible matrix. Adopting TICA for data pre-processing ensures that the mapping $(\mathbf z_t,\mathbf v_t)=F(\mathbf x_t^{\mathrm{TICA}})=F(\mathbf W(\mathbf x_t-\bar{\mathbf x}))$ given by RC flow is still invertible with respect to $\mathbf x_t$. We can then reconstruct the original configuration $\mathbf x_t$ from the RC $\mathbf z_t$ and noise $\mathbf v_t$ as
\begin{eqnarray*}
\mathbf{x}_{t} & = & \mathbf{W}^{-1}\mathbf{x}_{t}^{\mathrm{TICA}}+\bar{\mathbf{x}}\\
 & = & \mathbf{W}^{-1}F^{-1}\left(\mathbf{z}_{t},\mathbf{v}_{t}\right)+\bar{\mathbf{x}}.
\end{eqnarray*}

However, if $\mathbf x_t$ contains linearly dependent elements and $\mathbf C(0)$ is numerically singular, TICA removes the linear dependence and provides $\mathbf W$ with row number smaller than $D$. In this case, we can obtain an approximate inverse mapping from $(\mathbf z_t,\mathbf v_t)$ to $\mathbf x_t$ by minimizing the error $\left\Vert \delta\mathbf{x}_{t}-\widehat{\mathbf{W}^{-1}}\mathbf{x}_{t}^{\mathrm{TICA}}\right\Vert ^{2}=\left\Vert \delta\mathbf{x}_{t}-\widehat{\mathbf{W}^{-1}}\mathbf{W}\delta\mathbf{x}_{t}\right\Vert ^{2}$, which yields
\[
\widehat{\mathbf{W}^{-1}} = \delta\mathbf{X}\left(\mathbf{W}\delta\mathbf{X}\right)^{+}.
\]
Here $\delta \mathbf X=(\delta\mathbf x_1,\ldots,\delta\mathbf x_T)$ and the superscript $+$ denotes the Moore-Penrose pseudo inverse.
Finally, we can approximately reconstruct configuration $\mathbf x_t$ from $(\mathbf z_t,\mathbf v_t)$ by
\[
\mathbf x_t\approx \widehat{\mathbf W^{-1}}F^{-1}\left(\mathbf{z}_{t},\mathbf{v}_{t}\right)+\bar{\mathbf{x}}.
\]

\section{Transfer operators of $\{\mathbf x_t\}$ and $\{\mathbf z_t\}$}\label{app:transfer-operator}
We first briefly introduce properties of the transfer operator $\mathcal T_{\tau}^z$ of \eqref{eq:continuous-brownian}, which is defined by
\[
\mathcal{T}_{\tau}^{z}h(\mathbf{z})=\int\frac{\mu^{z}(\mathbf{z}')}{\mu^{z}(\mathbf{z})}p_{\tau}^{z}(\mathbf{z}',\mathbf{z})h(\mathbf{z}')\mathrm{d}\mathbf{z}',\text{ for }\left\langle h,h\right\rangle _{\mu^{z}}<\infty.
\]
For more details, we refer to \cite{klus2018data} and references therein.
Due to the time-reversibility of the Brownian dynamics, $\mathcal T_{\tau}^z$ is a self-adjoint operator
and can be written in terms of the eigenfunctions as
\[
\mathcal{T}_{\tau}^{z}h=\sum_{i=0}^{\infty}\lambda_i^\tau \left\langle h,r_i^z\right\rangle_{\mu^z}r_i^z
\]
Here
$1=\lambda_0^\tau>\lambda_1^\tau\ge \lambda_2^\tau\ge\ldots$ are eigenvalues,
which are associated with implied time scales as $t_i=-\tau/\log(\lambda_t^\tau)$,
$r_0^z=\mu^z,r_1^z,r_2^z,\ldots$ are normalized eigenfunctions with $\left\langle r_i^z,r_j^z\right\rangle_{\mu^z}=1_{i=j}$, and the inner product
$\left\langle h,h'\right\rangle_{\mu^z}=\int h(\mathbf z)h'(\mathbf z)\mu(\mathbf z)\mathrm d\mathbf z$.
Then, we have
\begin{eqnarray*}
p_{\tau}^{z}(\mathbf{z}',\mathbf{z}) & = & \frac{\mu^{z}(\mathbf{z})}{\mu^{z}(\mathbf{z}')}\mathcal{T}_{\tau}^{z}\delta_{\mathbf{z}'}(\mathbf{z})\\
 & = & \frac{\mu^{z}(\mathbf{z})}{\mu^{z}(\mathbf{z}')}\sum_{i=0}^{\infty}\lambda_{i}^{\tau}\mu^{z}(\mathbf{z}')r_{i}^{z}(\mathbf{z}')r_{i}^{z}(\mathbf{z})\\
 & = & \sum_{i=0}^{\infty}r_{i}^{z}(\mathbf{z}')\cdot\lambda_{i}^{\tau}r_{i}^{z}(\mathbf{z})\mu^{z}(\mathbf{z})
\end{eqnarray*}

Next, we prove that \eqref{eq:eigendecomposition-x} holds for RC flow.
According to the decomposition of $p_\tau^z$ and \eqref{eq:reconstruction-rcf},
\begin{eqnarray*}
p_{\tau}^{x}(\mathbf{x}',\mathbf{x}) & = & p_{\tau}^{z}(\mathbf{z}',\mathbf{z})S(\mathbf{x})\\
 & = & \sum_{i=0}^{\infty}r_{i}^{z}(\mathbf{z}')\cdot\lambda_{i}^{\tau}r_{i}^{z}(\mathbf{z})\mu^{z}(\mathbf{z})S(\mathbf{x})\\
 & = & \sum_{i=0}^{\infty}r_{i}^{x}(\mathbf{x}')\cdot\lambda_{i}^{\tau}r_{i}^{x}(\mathbf{x})\mu^{z}(\mathbf{x}),
\end{eqnarray*}
where $\ensuremath{F(\mathbf{x})=(\mathbf{z},\mathbf{v})}$, $F(\mathbf{x}')=(\mathbf{z}',\mathbf{v}')$
and $r_{i}^{x}(\mathbf{x})=r_{i}^{z}(\Phi(\mathbf{x}))$.

Moreover, based on the above analysis,
the eigendecomposition of the transfer operator of $\{\mathbf x_t\}$ can be written as
\begin{eqnarray*}
\mathcal{T}_{\tau}^{x}h(\mathbf{x}) & \triangleq & \int\frac{\mu^{x}(\mathbf{x}')}{\mu^{x}(\mathbf{x})}p_{\tau}^{x}(\mathbf{x}',\mathbf{x})h(\mathbf{x}')\mathrm{d}\mathbf{x}' \\
 & = & \sum_{i=0}^{\infty}\int r_{i}^{x}(\mathbf{x}')h(\mathbf{x}')\mu^{x}(\mathbf{x}')\mathrm{d}\mathbf{x}'\cdot\lambda_{i}^{\tau}r_{i}^{x}(\mathbf{x}) \\
 & = & \sum_{i=0}^{\infty}\lambda_{i}^{\tau}\left\langle h,r_{i}^{x}\right\rangle _{\mu^{x}}r_{i}^{x}(\mathbf{x}).
\end{eqnarray*}
By considering
\begin{eqnarray*}
\left\langle r_{i}^{x},r_{j}^{x}\right\rangle _{\mu^{x}} & = & \int r_{i}^{x}(\mathbf{x})r_{j}^{x}(\mathbf{x})\mu^{x}(\mathbf{x})\mathrm{d}\mathbf{x}\\
 & = & \iint r_{i}^{z}(\mathbf{z})r_{j}^{z}(\mathbf{z})\mu^{z}(\mathbf{z})\mathcal{N}(\mathbf{v}|\mathbf{0},\mathbf{I})\mathrm{d}\mathbf{z}\mathrm{d}\mathbf{v}\\
 & = & \left\langle r_{i}^{z},r_{j}^{z}\right\rangle _{\mu^{z}}\\
 & = & 1_{i=j},
\end{eqnarray*}
we can conclude that $\mathcal{T}_{\tau}^{x}r_i^x=\lambda_i^\tau r_i^x$, i.e., $\lambda_i^\tau$ and $r_i^x(\mathbf x)=r_i^z(\Phi(\mathbf x))$ are the $i$th eigenvalue and eigenfunction of $\mathcal{T}_{\tau}^{x}$.

\section{Proof of \eqref{eq:mi-bound}}\label{sec:mi-bound-proof}
We show here the derivation of \eqref{eq:mi-bound} for the self-completeness of the paper. The similar proof can be found in \cite{wang2019past}.

For convenience of notation, we denote by
\[
\hat{\mathbb{P}}(\mathbf{x}_{t+\tau}|\mathbf{z}_{t})=p_{\tau}^{z}(\mathbf{z}_{t},\Phi(\mathbf{x}_{t+\tau}))S(\mathbf{x}_{t+\tau})
\]
the approximation of the conditional distribution $\mathbb{P}(\mathbf{x}_{t+\tau}|\mathbf{z}_{t})$ given by RC flow. It can be seen from \eqref{eq:reconstruction-rcf} that $\hat{\mathbb{P}}(\mathbf{x}_{t+\tau}|\mathbf{z}_{t})=\hat p_\tau^x(\mathbf{x}_t,\mathbf{x}_{t+\tau})$ in RC flow.
According to the definition of mutual information, we have
\begin{eqnarray*}
\mathrm{MI}(\mathbf{z}_{t},\mathbf{x}_{t+\tau}) & = & \mathbb{E}\left[\log\mathbb{P}(\mathbf{x}_{t+\tau}|\mathbf{z}_{t})\right]+H(\mathbf{x}_{t+\tau})\\
 & = & \mathbb{E}\left[\log\frac{\mathbb{P}(\mathbf{x}_{t+\tau}|\mathbf{z}_{t})}{\hat{\mathbb{P}}(\mathbf{x}_{t+\tau}|\mathbf{z}_{t})}\right]+\mathbb{E}\left[\log\hat{\mathbb{P}}(\mathbf{x}_{t+\tau}|\mathbf{z}_{t})\right]\\
 & & +H(\mathbf{x}_{t+\tau})\\
 & = & \mathbb{E}_{\mathbf{z}_{t}}\left[\mathbb{E}_{\mathbf{x}_{t+\tau}\sim\mathbb{P}(\mathbf{x}_{t+\tau}|\mathbf{z}_{t})}\left[\log\frac{\mathbb{P}(\mathbf{x}_{t+\tau}|\mathbf{z}_{t})}{\hat{\mathbb{P}}(\mathbf{x}_{t+\tau}|\mathbf{z}_{t})}\right]|\mathbf{z}_{t}\right]\\
 &  & +\mathbb{E}\left[\log\hat{p}_{\tau}^{x}(\mathbf{x}_{t},\mathbf{x}_{t+\tau})\right]+H(\mathbf{x}_{t+\tau})
\end{eqnarray*}
where $H$ denotes the entropy and $\mathbb E$ denotes the mean value over all transition pairs $(\mathbf x_t,\mathbf x_{t+\tau})$ in trajectories. Notice that $\mathbb{E}_{\mathbf{x}_{t+\tau}\sim\mathbb{P}(\mathbf{x}_{t+\tau}|\mathbf{z}_{t})}\left[\log\frac{\mathbb{P}(\mathbf{x}_{t+\tau}|\mathbf{z}_{t})}{\hat{\mathbb{P}}(\mathbf{x}_{t+\tau}|\mathbf{z}_{t})}\right]$ equals the KL divergence between $\mathbb{P}(\mathbf{x}_{t+\tau}|\mathbf{z}_{t})$ and $\hat{\mathbb{P}}(\mathbf{x}_{t+\tau}|\mathbf{z}_{t})$, which is always non-negative. Therefore, in the limit case of infinite data size, we can obtain
\[
\mathbb{E}\left[\log\hat{\mathbb{P}}(\mathbf{x}_{t+\tau}|\mathbf{z}_{t})\right]=-\mathcal L_{\mathrm{kin}}
\]
and
\[
\mathrm{MI}(\mathbf{z}_{t},\mathbf{x}_{t+\tau})\ge-\mathcal{L}_{\mathrm{kin}}+H(\mathbf{x}_{t+\tau}),
\]
where the last term is independent of parameters of RC flow. The equality holds only if $\hat{\mathbb{P}}(\mathbf{x}_{t+\tau}|\mathbf{z}_{t})=\mathbb{P}(\mathbf{x}_{t+\tau}|\mathbf{z}_{t})$.

\section{Calculation of $p^z_{\tau}$}\label{app:pz}
In the importance sampling method \cite{durham2002numerical}, the time interval $[t,t+\tau]$ is divided into $M$ sub-intervals of length $\Delta=\tau/M$.
Letting $\mathbf u_{0}=\mathbf z_{t}$, $\mathbf u_{1}=\mathbf z_{t+\Delta}$, ..., $\mathbf u_{M}=\mathbf z_{t+M\Delta}=\mathbf z_{t+\tau}$ and applying Euler-Maruyama discretization to each sub-interval, we have
\begin{eqnarray}
p_{\tau}^{z}(\mathbf{z}_{t},\mathbf{z}_{t+\tau}) & = & \int\mathbb{P}(\mathbf{u}_{1}|\mathbf{u}_{0})\ldots\mathbb{P}(\mathbf{u}_{M}|\mathbf{u}_{M-1})\mathrm{d}\mathbf{u}_{1:M-1}\nonumber\\
 & = & \int f(\mathbf{u}_{0},\mathbf{u}_{1})\ldots f(\mathbf{u}_{M-1},\mathbf{u}_{M})\mathrm{d}\mathbf{u}_{1:M-1}\label{eq:pz-integral},
\end{eqnarray}
where $u_{1:M-1}=(u_{1},\ldots,u_{M-1})$ and
\[
f(\mathbf{u},\mathbf{u}')=\mathcal{N}\left(\mathbf{u}'|\mathbf{u}-\nabla V(\mathbf{u})\cdot\Delta,2\Delta\mathbf{I}\right).
\]
According to \cite{durham2002numerical}, We can draw $K_s$ samples of $\mathbf u_{1:M-1}$ from the proposal density
\[
\mathbf{u}_{1:M-1}^{k}\sim\prod_{m=0}^{M-2}g_{m}(\mathbf{u}_{m}^{k},\mathbf{u}_{m+1}^{k}),\quad\text{for }k=1,\ldots,K_s
\]
and calculate the integral \eqref{eq:pz-integral} by importance sampling
\begin{equation}\label{eq:pz-is}
p_{\tau}^{z}(\mathbf{z}_{t},\mathbf{z}_{t+\tau})\approx\frac{1}{K_s}\sum_{k=1}^{K_s}\frac{\prod_{m=0}^{M-1}f(\mathbf{u}_{m}^{k},\mathbf{u}_{m+1}^{k})}{\prod_{m=0}^{M-2}g_{m}(\mathbf{u}_{m}^{k},\mathbf{u}_{m+1}^{k})}
\end{equation}
where $\mathbf{u}_{0}^{k}\equiv\mathbf u_0=\mathbf z_t$, $\mathbf{u}_{M}^{k}\equiv\mathbf u_M=\mathbf z_{t+\tau}$, and
\begin{equation}\label{eq:g}
g_{m}(\mathbf{u},\mathbf{u}')=\mathcal{N}\left(\mathbf{u}'|\mathbf{u}+\frac{\mathbf{u}_{M}-\mathbf{u}}{M-m},\frac{2\Delta\left(M-m-1\right)}{M-m}\mathbf{I}\right).
\end{equation}

The whole procedure of the approximation of $p_\tau^z$ is summarized in Algorithm \ref{alg:pz}.
\begin{algorithm}
  \caption{Importance sampling approximation of $p_\tau^z(\mathbf z_t,\mathbf z_{t+\tau})$}
  \label{alg:pz}
  \SetAlgoNoLine
   \For{$k=1,\ldots,K_s$}{
       \For{$m=1,\ldots,M-1$}{
            Draw $\mathbf u_m^k\sim g_m(\mathbf u_{m-1}^k,\cdot)$ with \eqref{eq:g}.
       }
   }
   Calculate $p_\tau^z(\mathbf z_t,\mathbf z_{t+\tau})$ by \eqref{eq:pz-is}.
\end{algorithm}

\section{Model and implementation details}\label{app:details}
\subsection{Structure and hyperparameters of RC flow}
In this work, the RealNVP model of $F$ is implemented in the Bgflow package \footnote{{h}ttps://github.com/noegroup/bgflow}. It consists of $12$ affine coupling blocks, where shift and scale transformations are performed by multiple layer perceptrons (MLPs) with $3$ hidden layers of width $128$. In the GMM \eqref{eq:gmm} of $\mu^z$, $w(\mathbf c)$ and $\sigma(\mathbf c)$ are modeled by MLPs with $3$ hidden layers of width $64$, and $K=40$.
The weight $\alpha$ in \eqref{eq:loss} is set to be $0.1$.

In Algorithm \ref{alg:rc}, Lines 6, 10 and 12 are implemented by Adam algorithm \cite{kingma2014adam}. Models are trained for $5$ epochs (Lines 6 and 10) and $20$ epochs (Line 12), and the learning rate is initially $10^{-3}$. Moreover, the learning rate is decayed by a factor of $0.1$ for every $5$ epochs when solving the optimization problem in Line 12. In Algorithm \ref{alg:pz}, $K_s=20$ and $M=10$.

For all our experiments, we implemented pre-training steps. Furthermore, we applied TICA transformations specifically to the examples shown in Figs.~\ref{fig:toy-example-part2} and \ref{fig:alanine} with a lag time of $10$ steps.

\subsection{Simulations}\label{app:simulations}
Potential functions of examples shown in Fig.~\ref{fig:toy-example-part1} are
\begin{eqnarray*}
V_{\text{double well}}(x_{1},x_{2}) & = & 5\left(x_{1}^{2}-1\right)^{2}+10\left(x_{1}^{2}+x_{2}-1\right)^{2},\\
V_{\text{Mueller}}(x_{1},x_{2}) & = & \sum_{i=1}^{4}A_{i}\exp\left(a_{i}\left(x_{1}-\bar{x}_{i}\right)^{2}+\right.\\
 &  & \left.b_{i}\left(x_{1}-\bar{x}_{i}\right)\left(x_{2}-\bar{y}_{i}\right)+c_{i}\left(x_{2}-\bar{y}_{i}\right)^{2}\right),
\end{eqnarray*}
where $(A_{1},\,$$\ldots,\,$$A_{4})$ $=$ $(-\frac{20}{3},\,$$-\frac{10}{3},\,$$-\frac{17}{3},\,$$\frac{1}{2})$, $(a_{1},\,$$\ldots,\,$$a_{4})$ $=$ $(-1,\,$$-1,\,$$-6.5,\,$$0.7)$,
$(b_{1},\,$$\ldots,\,$$b_{4})$ $=$ $(0,\,$$0,\,$$11,\,$$0.6)$,
$(c_{1},\,$$\ldots,\,$$c_{4})$ $=$ $(-10,\,$$-10,\,$$-6.5,\,$$0.7)$,
$(\bar x_{1},\,$$\ldots,\,$$\bar x_{4})$ $=$ $(1,\,$$0,\,$$-0.5,\,$$-1)$ and
$(\bar y_{1},\,$$\ldots,\,$$\bar y_{4})$ $=$ $(0,\,$$0.5,\,$$1.5,\,$$1)$. 
For each potential, we generate a trajectory containing $1.5\times 10^5$ frames by Euler-Maruyama discretization of the Brownian dynamics with the inverse temperature $0.5$ (double well potential) or $1$ (Mueller potential).
The time intervals between frames and step sizes of the discretization are $0.01, 2\times 10^{-4}$ (double well potential) and $0.025, 5\times 10^{-4}$ (Mueller potential).

For the example illustrated by Fig.~\ref{fig:toy-example-part2},
$V'(s_1,s_2)=\cos(7\cdot\mathrm{atan2}(s_2,s_1))$. The trajectory of $\mathbf s$ containing $3\times 10^5$ frames is also generated by the Euler-Maruyama scheme, where the inverse temperature is $1$, the time interval between frames is $0.01$ and the discretization step size is $2\times 10^{-4}$. The simulation data in the space of $\mathbf x$ are obtained via the transformation
\begin{eqnarray}
x_{1} & = & s_{1}^{\prime}\cos s_{1}^{\prime}+\frac{u_{1}}{\sqrt{u_{1}^{2}+u_{3}^{2}}}s_{3},\nonumber\\
x_{2} & = & s_{2}^{\prime},\nonumber\\
x_{3} & = & s_{1}^{\prime}\sin s_{1}^{\prime}+\frac{u_{3}}{\sqrt{u_{1}^{2}+u_{3}^{2}}}s_{3},\label{eq:swiss-transform}
\end{eqnarray}
where
$u_{1} = \sin s_{1}^{\prime}+s_{1}^{\prime}\cos s_{1}^{\prime}$,
$u_{3} = -\cos s_{1}^{\prime}+s_{1}^{\prime}\sin s_{1}^{\prime}$,
$s_{1}^{\prime} = 3\pi\left(s_{1}+4\right)/4$ and
$s_{2}^{\prime} = 3\pi\left(s_{2}+4\right)/4$.

\subsection{Calculation of implied time scales}\label{app:its}
In our examples, implied time scales are all calculated by $50$-state Markov state models built by \texttt{pyEMMA}, where the spatial discretization is performed by k-means clustering.
Implied time scales of full-state kinetics are obtained from simulation trajectories in the space of $\mathbf x$ (two well and Mueller potentials), $\mathbf s$ (Swiss roll), and $(\phi,\psi)$ (alanine dipeptide). For reduced kinetics defined by \eqref{eq:continuous-brownian}, implied time scales
are obtained from
trajectories with the same sizes as training data,
which are also generated by the Euler-Maruyama discretization of \eqref{eq:continuous-brownian}.

\bibliographystyle{aipnum4-1}
\bibliography{rcf}

\end{document}